\documentclass[conference]{IEEEtran}
\IEEEoverridecommandlockouts
\usepackage{cite}
\usepackage{amsmath,amssymb,amsfonts}
\usepackage{graphicx}
\usepackage{textcomp}
\usepackage{xcolor}
\usepackage{algorithm}
\usepackage{algorithmic}
\usepackage{comment}
\usepackage{relsize}

\def\BibTeX{{\rm B\kern-.05em{\sc i\kern-.025em b}\kern-.08em
    T\kern-.1667em\lower.7ex\hbox{E}\kern-.125emX}}

\begin{document}

\title{Global Outlier Detection in a Federated Learning Setting with Isolation Forest\\
\thanks{This work was supported in part by Innosuisse through the Innosuisse Flagship project SPEARHEAD (PFFS-21-15).}
}

\author{\IEEEauthorblockN{1\textsuperscript{st} Daniele Malpetti}
\IEEEauthorblockA{%\textit{dept. name of organization (of Aff.)} \\
\textit{IDSIA (USI-SUPSI)}\\
Lugano, Switzerland \\
daniele.malpetti@idsia.ch}
\and
\IEEEauthorblockN{2\textsuperscript{nd} Laura Azzimonti}
\IEEEauthorblockA{%\textit{dept. name of organization of Aff.)} \\
\textit{IDSIA (USI-SUPSI)}\\
Lugano, Switzerland \\
laura.azzimonti@idsia.ch}
}

\maketitle

\begin{abstract}
We present a novel strategy for detecting global outliers in a federated learning setting, targeting in particular cross-silo scenarios. Our approach involves the use of two servers and the transmission of masked local data from clients to one of the servers. The masking of the data prevents the disclosure of sensitive information while still permitting the identification of outliers. Moreover, to further safeguard privacy, a permutation mechanism is implemented so that the server does not know which client owns any masked data point. The server performs outlier detection on the masked data, using either Isolation Forest or its extended version, and then communicates outlier information back to the clients, allowing them to identify and remove outliers in their local datasets before starting any subsequent federated model training. This approach provides comparable results to a centralized execution of Isolation Forest algorithms on plain data.
\end{abstract}


\section{Introduction}

Federated learning (FL) is a machine learning paradigm where multiple parties collaborate to train a shared machine learning model without centralizing data at a single location \cite{mcmahan2017communication}. During model training, data holders refrain from directly exchanging raw data; instead, they share model parameters such as gradients, weights, or other forms of processed information. This distributed learning paradigm is typically facilitated by a coordinating server, often referred to as the \textit{aggregator}, which collects local contributions from data holders, commonly known as \textit{clients}, and aggregates them to create a global model. Model training may adopt iterative schemes where an updated global model is sent to the clients at each iteration.

The applicability of FL spans diverse contexts, each driven by distinct needs. For instance, one common scenario involves a limited number of data-holding entities collaborating to train a global model without directly sharing their data, often for privacy reasons. This situation is known as the \textit{cross-silo} scenario and is notably observed in highly regulated domains like biomedicine, where FL is expected to become a prevalent technology \cite{rieke2019npj}. Conversely, another scenario involves a multitude of edge devices acting as clients. This is referred to as the \textit{cross-device} scenario and is common in IoT deployments. In this case, FL primarily aims to reduce the time and cost associated with centralized data transfer, while also addressing privacy needs.

It is worth noting that even though FL permits data owners to retain sovereignty over the usage of their own data,  it does not inherently guarantee security. In several cases, information can be reconstructed about the data used in training from models or model parameters \cite{geiping2020inverting, haim2022reconstructing}. Therefore, to ensure further privacy and security, various techniques such homomorphic encryption \cite{gentry2009fully}, secure multiparty computation protocols \cite{zhao2019secure}, or differential privacy \cite{dwork2006differential} are commonly employed. 

Similar to other machine learning models, FL training is susceptible to outliers or anomalies in data, which can detrimentally impact model performance. Furthermore, in a federated setting, outliers can be classified as \textit{local} outliers, which are outliers for a given client, and \textit{global} outliers, which are outliers overall. Across several domains, it is common to find examples of data points that are local outliers but not global outliers. For example, in the medical field, a given medical condition may be common in one region and rare in another \cite{ford2005geographic}. Therefore, in a study conducted at a center located in a low-prevalence region, individuals suffering from that condition may appear as local outliers. However, if the center participates in a FL multicenter study including centers in areas where the condition is more common, those individuals would not appear as global outliers. In most cases, for the training of FL models, a consortium would be interested in discarding global outliers and retaining local ones.

In centralized environments, various strategies have been developed over the years for outlier detection using a wide range of techniques \cite{chandola2009anomaly, ruff2021unifying}. These include statistical methods like z-score and modified z-score, distance-based algorithms such as k-nearest neighbors, density-based approaches like Local Outlier Factor (LOF) \cite{breunig2000lof}, tree-based models like Isolation Forest \cite{liu2008isolation} and its variants, as well as deep learning approaches \cite{pang2021deep}.

Despite the ubiquity of outlier detection, only a limited number of solutions tailored for federated contexts exist. Furthermore, such solutions predominantly focus on the IoT \textit{cross-device} setting, where anomaly detection is intrinsic, typically signaling device malfunctions or intrusions in IoT networks. Few solutions have been specifically designed for the \textit{cross-silo} setting, where outlier detection serves mainly as a preprocessing step aimed at identifying and removing outliers to enhance the quality of subsequently trained FL models.

In this article, we introduce a methodology focused on outlier detection within a FL framework, using the Isolation Forest (IF) algorithm \cite{liu2008isolation} or its Extended Isolation Forest (EIF) \cite{hariri2019extended} variant. The method is designed for a \textit{cross-silo} scenario, where two servers are present, and where clients hold data described by the same variables (i.e., horizontally partitioned data). In our approach, the principal server receives a masked version of the data that preserves the ``isolationness" of outliers, conducts outlier detection on these masked data, and communicates results to the clients. Notably, thanks to a permutation procedure operated with the help of the auxiliary server, the principal server does not know to which clients the identified outliers belong. 

The article is structured as follows: Section \ref{sec:related} examines current methods for outlier detection in a federated context, Section \ref{sec:background} provides a concise overview of the main algorithms and techniques used in our solution, Section \ref{sec:methodology} presents our methodology, and Section \ref{sec:experiments} outlines the experiments conducted and presents the results. In Section \ref{sec:discussion}, we delve into key aspects of our solution, particularly focusing on its security implications. Finally, Section \ref{sec:conclusion} offers concluding remarks and discusses potential extensions of the method to other contexts.

\section{Related work} \label{sec:related}

Federated outlier detection is increasingly leveraged in Internet of Things (IoT) systems in order to ensure reliable and timely identification of anomalies, while maintaining data privacy and network efficiency. Indeed, a federated approach not only preserves the privacy of individual devices but also reduces the need for extensive data transfer, minimizing latency and bandwidth usage. In particular, federated deep learning techniques, based, e.g., on long short-term memory (LSTM), gated recurrent units (GRUs) and convolutional neural networks (CNNs), have been recently proposed to predict intrusions in IoT networks \cite{mothukuri, 8884802, 9195012} or device failures \cite{9146846}.
A recent work, also targeted at the IoT domain, implements a federated version of IF \cite{li2023federated}, where a global isolation tree is built from local encrypted contributions after the use of differential privacy locally. 

As observed in the introduction, outside of the IoT domain, there are very few examples of methods specifically developed for cross-silo scenarios, where outlier detection mainly constitutes a preprocessing step for the subsequent training of a FL model. A notable example is  a privacy-preserving version of LOF \cite{li2015privacy}, which was developed prior to the introduction of the term FL.

\section{Background} \label{sec:background}

In this section, we provide a brief overview of the main characteristics of both IF and EIF, as well as a short review of the homomorphic encryption Paillier cryptosystem, which we use a few times in our work. For an in-depth analysis and comparison of the IF and EIF, we refer the reader to \cite{chabchoub2022depth}. It is worth mentioning that, in addition to EIF, there exist several other extensions or improvements of the IF algorithm, such as SA-IForest \cite{xu2017improved}, E-IForest \cite{liao2019entropy}, and LSHIForest \cite{zhang2017lshiforest}. For the sake of simplicity, in this work, we decided to focus on the original algorithm and on its most well-known extension, as they both remain widely used.

\subsection{Isolation Forest}

The IF algorithm is based on the principle that anomalies (outliers) are easier to isolate than normal data points (inliers). The algorithm consists of two different phases: a training phase, which builds the  forest, and a scoring phase, which assigns each data point an outlier score.

\begin{figure}[t]%[htbp] 
\label{fig:pictorial}
\centerline{\includegraphics[scale=.41]{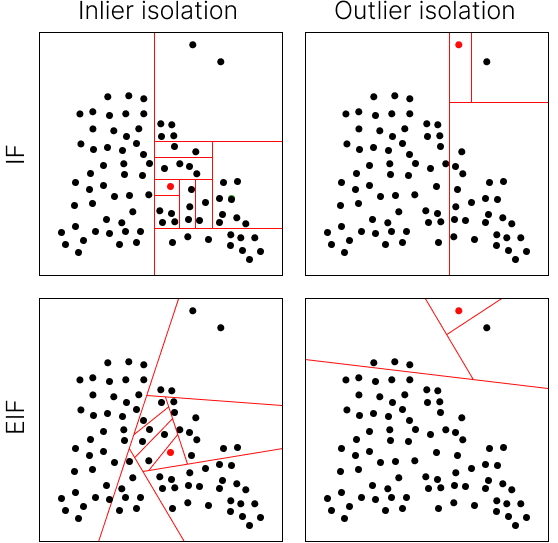}}
\caption{Pictorial representation of the steps conducted by IF and EIF to isolate an intlier and an outlier in a two-dimensional dataset.}
\end{figure}

During the training phase, a forest of $t$ binary trees is created, with each tree using a different set of $\psi$ data points randomly selected from the entire dataset. For every tree, the algorithm starts with all the $\psi$ data points in a root node, and then randomly chooses a direction for splitting (i.e., horizontal or vertical), creating a split within the range defined by the minimum and maximum values assumed by the data points. This creates two child nodes in the tree, where the data points are stored. The same procedure is repeated iteratively for each of the nodes. The splitting of a given node stops if there is a single data point in the node or if a maximum tree depth parameter ($\log_2 (\psi)$) is reached. Nodes without any child nodes are called \textit{external nodes}, whereas the others are called \textit{internal nodes}. Figure~\ref{fig:pictorial} shows an example of several splits leading to the isolation (i.e., the storing in an external node) of an inlier and an outlier respectively. It is clear that the number of splits required to isolate the outlier is much smaller than the number of splits required to isolate the inlier.

During the scoring phase, every tree in the forest evaluates each data point in the dataset, assigning each data point to an external node within that tree. Subsequently, the distance (path length) between the external node containing the data point and the root node is computed for each data point and for each tree. These distances are then averaged across all trees for each data point, yielding an average path length that characterizes each data point. Outlier scores are then calculated based on the average path lengths, with shorter average path lengths resulting in higher outlier scores.

\subsection{Extended Isolation Forest}

There exist scenarios where the IF algorithm fails to produce satisfactory results, particularly when the data has symmetries like rotational symmetries, which are not reflected in the outlier scores. To address these limitations, the Extended Isolation Forest (EIF) was developed. The difference between IF and EIF resides in the training phase, whereas the scoring phase is identical. In EIF, the splits are not performed by means of horizontal or vertical hyperplanes, but by  selecting a random hyperplane from the set of all possible ones (see Figure~\ref{fig:pictorial}). 
It is worth noting that, despite the more general approach, EIF does not systematically outperform IF \cite{chabchoub2022depth}.

\subsection{Paillier cryptosystem}

We briefly recall the main properties of a cryptosystem, with a focus on the Paillier cryptosystem. Every cryptosystem comprises three fundamental algorithms:
\begin{itemize}
    \item The key generation algorithm $(s k, p k)=\operatorname{Gen}(keysize)$, which generates a secret key $s k$ and a public key $p k$ based on a security parameter $keysize$.
    \item The encryption algorithm $[x]_{pk}=\operatorname{Enc}(x, p k)$, which maps a plaintext $x$ to a ciphertext $[x]_{pk}$ using the public key $p k$.
    \item The decryption algorithm $x=\operatorname{Dec}([x]_{pk}, s k)$, which reverses the encryption process using the secret key $sk$ to recover the original plaintext $x$.
\end{itemize}
Please, note that in the following we will always explicitly indicate the key used for encrypting a given plaintext. This is essential because our pseudocodes involve the use of multiple key pairs, and not indicating the key could lead to ambiguities.

The Paillier cryptosystem constitutes an additive partially homomorphic cryptosystem. Let $x_1$ and $x_2$ be two plaintexts. Homomorphic addition is achieved through the product of two ciphertexts, namely it is defined as $\left[x_1\right]_{pk} \oplus \left[x_2\right]_{pk} := \left[x_1\right]_{pk} \cdot \left[x_2\right]_{pk}$, and ensures that
$
\left[x_1\right]_{pk} \oplus\left[x_2\right]_{pk} =\left[x_1 + x_2 \right]_{pk}
$.
Moreover, homomorphic multiplication by a plaintext is achieved through exponentiation, is defined as $x_1 \otimes \left[x_2\right]_{pk} := \left[x_2\right]_{pk}^{x_1}$, and ensures that
$
x_1 \otimes\left[x_2\right]_{pk} =  \left[x_1 x_2\right]_{pk} 
$.

\section{Methodology} \label{sec:methodology}

We propose a methodology centered around two servers: the principal server $\mathcal{P}$ and the auxiliary server $\mathcal{A}$. The distinction between these servers lies in the amount of information to which they have access as well as in their roles. The principal server $\mathcal{P}$ receives a masked version of client data and carries out the main task of outlier identification, and communicates results to clients. Through the knowledge of outlier scores, this server is aware of the presence and extent of outliers in the data. Conversely, the auxiliary server $\mathcal{A}$ only receives noise or encrypted information and does not have visibility on any information regarding the data apart from knowing the total number of data points involved in the process. Its presence primarily has the role of enhancing the security of the process. Alongside these servers, there are $m$ clients, indexed from 0 to $m-1$, 
with the $i$-th client denoted as $\mathcal{C}^i$. In the process, the two servers communicate with all the clients and between themselves, whereas no direct communication channel among clients is in place.  For this reason, the indexing of clients is purely a mathematical notation, not associated with any actual order among of them: client $\mathcal{C}^i$ does not know the identity of client $\mathcal{C}^{i-1}$ nor of client $\mathcal{C}^{i+1}$, and each of the two servers uses its own indexing system.

We operate under the assumption that both servers act as semi-honest parties, adhering to the working protocol without collusion or malicious intent to disrupt or poison the data. However, they may attempt to infer information about the original data from the information that they receive during the process.  Regarding the clients, we assume their adherence to the protocol without attempts to undermine it, but we can partially relax the assumption of non-collusion, as elaborated in Section \ref{sec:discussion}.

Our methodology consists of several preliminary steps followed by a main process. The preliminary steps focus on generating a set of integer values that will govern the main process phase, and on generating a set of matrices, mainly used to transform or mask the data. All of these operations are based exclusively on the use of randomly chosen values or metadata. In the main process, the actual data processing occurs, including the task of outlier detection. The full workflow is summarized in Algorithm \ref{algo:full}, where references to the subsections describing the specific steps are also provided.

In this section, we use $\delta_{ij}$ to denote the Kronecker delta and the shorthand notation 
$\mathbb{Z}_k = \{0, 1, 2, \ldots, k-1\}$ to denote the set of integers modulo $k$. The notation $A = \{A_i\}_{i \in \mathbb{Z}_k}$ represents an ordered set of $k$ elements, indexed from $0$ to $k-1$, where $A_i$ is the $i$-th element of the set, and $|A|$ is the cardinality of the set. Additionally, $W_{j\cdot}$ denotes the $j$-th row of a matrix $W$, with rows indexed starting from $0$ rather than $1$. Note that, for the sake of clarity, we use apices only for indices associated to clients: e.g. $N^i$ denotes the local sample size of client $\mathcal C^i$.

\subsection{Preliminary steps -- integers generation} \label{subsec:integers}

Initially, each client $\mathcal C^i$ generates a key pair $(pk^i,sk^i)=\operatorname{Gen}(keysize)$ within Paillier cryptosystem, shares the public key $pk^i$ with all the members of the consortium, and keeps the secret key $sk^i$ private. Please note that even though the Paillier cryptosystem is a key element of the method, it is used only a limited number of times and on a limited number of integers in order to minimize the computational overhead.

Next, the clients collectively agree on an integer number $\Xi$, unknown to the two servers. This integer is of great importance as it will serve as a global seed for conducting operations that involve random generations, ensuring consistency across all clients.  The consensus on $\Xi$ is achieved through Algorithm \ref{algo:clientclient}, based on the Paillier cryptosystem, which clients execute together with the auxiliary server $\mathcal A$. 

Subsequently, the clients, together with either of the two servers, jointly calculate the total number of data points involved in the process, $N$, which then becomes public to all parties. It is worth noting that our methodology does not require any server to know the individual number of data points held by each client; only the total number of data points is necessary for the method to function. This is a key aspect of the approach, as knowing the number of data points a client holds before and after the process would reveal how many outliers the client discarded. This step can be executed using any secure sum protocol according to the preference of the parties involved. Algorithm \ref{algo:clientclient} can also be used for this purpose by using the local sample sizes $\{N^i\}_{i \in \mathbb{Z}_m}$ as inputs instead of randomly generated numbers, and it does so without the need of establishing direct communication among clients. 

In the subsequent step, clients collaborate with the auxiliary server $\mathcal{A}$ so that each client $\mathcal{C}^i$ is assigned a set of non-consecutive integers, denoted as $\tilde Z^i$, which satisfies  $|\tilde{Z}^i| = N^i$ $\forall i \in \mathbb{Z}_m$, $\tilde{Z}^i \cap \tilde{Z}^j=\varnothing$ $\forall i, j \in \mathbb{Z}_m$ such that $i\neq j$, and $\cup_{i \in \mathbb{Z}_m} \tilde{Z}^i = \mathbb{Z}_N$. This is achieved following Algorithm \ref{algo:starting}, which also makes use of the Paillier cryptosystem. The algorithm first assigns to each client $\mathcal{C}^i$ a starting point $s^i$, such that $s^i+N^i \equiv s^{i+1} \pmod N$, with $\mathcal{A}$ introducing an offset $H>0$ so that $s^0 \neq 0$. Please recall that, as already observed, $\mathcal{C}^i$ does not know the identity of the client who receives $s^{i+1}$, as the indexing is only known to $\mathcal{A}$.  Then, each client permutes the full set of integers in $\mathbb{Z}_N$ using a previously agreed-upon permutation function, with the global seed $\Xi$, ensuring that they all permute the numbers in the same way. Each client $\mathcal{C}^i$ selects the integers in positions 
$\{s^i \mod N,...,(s^i+N^i-1)  \mod N\}$ in the permuted set, thus creating the sets $\tilde{Z}^i$. These sets consist of non-consecutive integers, are non-intersecting, and their union covers the entire set of integers $\mathbb{Z}_N$. The sets $\tilde{Z}^i$ will be used by clients to conduct operations on different rows of an $N$-row matrix, ensuring both that a client does not use a block of consecutive rows and that each client uses distinct rows.

At the conclusion of these steps, the servers have knowledge of $N$, and the $i$-th client $\mathcal{C}^i$ of $N$, $\Xi$, $\tilde{Z}^i$.

\begin{algorithm}[t]
\caption{Client-client integer agreement protocol}
\label{algo:clientclient}
\textbf{Input:} Public keys $\{pk^i\}_{i \in \mathbb Z_m}$ held by all parties. \\
\textbf{Output:} Integer $\Xi$ held by all clients. \\
\textbf{Procedure:}
\begin{algorithmic}[1]
    \FOR{$i \in \mathbb{Z}_m$ \textbf{in parallel}}
        \STATE $\mathcal{C}^i$ randomly generates an integer $\xi^i$
        \STATE $\mathcal{C}^i$ encrypts $\{[\xi^i]_{pk^j} = \operatorname{Enc} (\xi^i, {pk^j})\}_{j \in \mathbb Z_m}$
        \STATE $\mathcal{C}^i$ sends $\{[\xi^i]_{pk^j}\}_{j \in \mathbb Z_m}$ to $\mathcal{A}$
    \ENDFOR 
    \FOR{$i \in \mathbb{Z}_m$ \textbf{in parallel}}
         \STATE $\mathcal{A}$ calculates $ 
         [\Xi]_{pk^i} = \mathlarger{\mathlarger{{\mathlarger{\oplus}}}}_{j \in \mathbb Z_m} [\xi^j]_{pk^i}$
         \STATE $\mathcal{A}$ sends $[\Xi]_{pk^i}$ to $\mathcal{C}^i$
        \STATE $\mathcal{C}^i$ decrypts $\Xi = \operatorname{Dec} ([\Xi]_{pk^i}, sk^i)$
    \ENDFOR
\end{algorithmic}
\end{algorithm}

\begin{algorithm}[t]
\caption{Indices assignment}
\label{algo:starting}
\textbf{Input:} Public keys $\{pk^i\}_{i \in \mathbb Z_m}$ held by all parties. Local sample size $N^i$ held by $\mathcal C^i$. Global seed $\Xi$ held by all clients. $\{[\xi^i]_{pk^j}\}_{i,j \in \mathbb Z_m}$ held by $\mathcal A$ from a previous execution of Algorithm \ref{algo:clientclient}.\\
\textbf{Output:} Sets $\{  \tilde{Z}^i \}_{i \in \mathbb Z_m}$, with $Z^i$ held by $\mathcal C^i$. \\
\textbf{Procedure:} 
\begin{algorithmic}[1]
\FOR{$i \in \mathbb{Z}_m$ \textbf{in parallel}}
    \STATE $\mathcal{C}^i$ encrypts $\{  [N^i]_{pk^j } =  \operatorname{Enc} (N^i,pk^j) \}_{j \in \mathbb Z_m} $
    \STATE $\mathcal{C}^i$ sends $ \{  [N^i]_{pk^j }\}_{j \in \mathbb Z_m}$ to $\mathcal A$
\ENDFOR
\STATE $\mathcal{A}$ generates a random integer  $h \in \mathbb Z_m$
\FOR{$i \in \mathbb{Z}_m$ \textbf{in parallel}}
    \STATE $\mathcal{A}$ calculates $[H]_{pk^i} = \mathlarger{\mathlarger{{\mathlarger{\oplus}}}}_{j \in \mathbb Z_h} [\xi^j]_{pk^i}$
    \STATE $\mathcal A$ calculates ${[s^{i}]_{pk^{i}} = [H]_{pk^i} \oplus
    (1-\delta_{0i}) \mathlarger{\mathlarger{{\mathlarger{\oplus}}}}_{j \in \mathbb Z_{i-1} } [N^j]_{pk^i }}$
    \STATE $\mathcal A$ sends $[s^i]_{pk^i}$ to $\mathcal C^i$
    \STATE $\mathcal{C}^i$ decrypts $s^i = \operatorname{Dec}( [s^i]_{pk^{i}}, sk^i ) $
    \STATE $\mathcal{C}^i$ calculates $Z=\operatorname{permute}(\mathbb Z_N, \, \Xi)$
    \STATE $\mathcal{C}^i$ sets $\tilde{Z}^i=\{ Z_j \}_{j \in \{s^i \mod N, ..., (s^i+N^i-1) \mod N \}}$
\ENDFOR
\end{algorithmic}
\end{algorithm}

\subsection{Preliminary steps -- matrices generation}

All clients generate the same real invertible matrix $M$ locally, out of the product of three matrices. Specifically, each client uses the integer $\Xi$ as a seed to generate both an orthogonal matrix $Q$ and a diagonal invertible matrix $S$, and the seed $\Xi+1$ to generate another orthogonal matrix $Q'$. The matrix $M$ is then calculated as $M=QSQ'$. Various tools are available for generating orthogonal matrices, such as the \texttt{pracma} \cite{pracma}  package in R and the \texttt{scipy.stats} module in Python. For the invertible diagonal matrix $S$, values are generated uniformly in the interval $(1, T)$, where $T$ is a specified threshold greater than 1. The lower bound 1 is chosen so that the matrix is numerically far from singularity. We call $M$ the masking transformation, as clients will use it as a multiplicative mask for their data.

Next, each client creates a noise matrix $R^i$ locally, with as many rows as the total number of data points $N$ and as many columns as the number of variables $D$. Clients will sum their data points to rows of $R^i$, using it as an additive mask. To create $R^i$, clients sample matrix elements independently from a Gaussian distribution $\mathcal{N}(0, \sigma^2)$, with $\sigma^2=10^{12}$. In this, we follow the same approach as in \cite{nasirigerdeh2022splink}. There, the authors use a framework \cite{nasirigerdeh2021hyfed} with a two-server topology analogous to the one presented in this article and use Gaussian noise to mask data too. As the authors mention, the choice of $\sigma^2=10^{12}$ should be reasonable in a variety of situations and could be modified according to specific needs.

At the conclusion of these steps, the $i$-th client $\mathcal{C}^i$ has generated the matrices $M$ and $R^i$.

\subsection{Main process -- data masking and transfer}

Once all the preliminary steps have been completed, the parties can start the main process, which makes use of all the integers and matrices generated in the preliminary steps.

As a first step, each client multiplies their local data $X^i$ by the matrix $M$, obtaining $\tilde{X}^i = X^i M$. It is important to note that the matrices are applied to the right of $X^i$, since data points are described by rows of the data matrices. Then, each client creates a matrix $W^i$ by summing the rows of $X^i$ to specific rows of $R^i$, namely to those with indices corresponding to the integers in $\tilde{Z}^i$. As clear from the observations in Section \ref{subsec:integers}, each client sums their own data to different rows of their noise matrix. Then, clients send $R^i$ to the auxiliary server $\mathcal A$ and $W^i$ to the principal server $\mathcal P$. $\mathcal A$ calculates $R=\sum_{i \in \mathbb Z_m} R^i$, whereas $\mathcal P$  calculates $W=\sum_{i \in \mathbb Z_m} W^i$. 

At this point, the server $\mathcal  A$ sends the aggregated noise matrix $R$ to $\mathcal P$. The server $\mathcal P$ can denoise their data and obtain the masked data matrix $X_{\text{masked}}=W-R$. This is a matrix where each row corresponds to a data point, and all data points have undergone the transformation $M$ (which is unknown to $\mathcal P$). Notably, thanks to the permutation operated with Algorithm \ref{algo:starting}, $\mathcal P$ does not know to which client belongs any data point. Moreover, data points belonging to same client are (in general) not stored in adjacent rows of the matrix. 

\begin{algorithm}[t]
\caption{Overall process}
\label{algo:full}
\textbf{Input:} Data $X^i$ and local sample size $N^i$ held by $\mathcal C^i$. \\
\textbf{Output:} Outliers information held by clients. \\
\textbf{Procedure:}
\begin{algorithmic}[1]
\STATE \#\#\# Preliminary steps -- subsection $A$
\FOR{$i \in \mathbb{Z}_m$ \textbf{in parallel}}
    \STATE $\mathcal{C}^i$ generates $(pk^i,sk^i)=\operatorname{Gen}(keysize)$
    \STATE $\mathcal{C}^i$ sends $pk^i$ to $\mathcal A  $ (or $ \mathcal P $)
    \STATE $\mathcal A  $ (or $ \mathcal P $) sends $pk^i$ to the rest of the parties
\ENDFOR
\STATE Clients and $\mathcal A$ execute Algorithm \ref{algo:clientclient} to find $\Xi$
\STATE Clients and $\mathcal A$ (or $\mathcal P$) securely calculate $N=\sum_{i \in \mathbb Z_m} N^i$
\STATE Clients and $\mathcal A$  execute Algorithm \ref{algo:starting} to find $\{\tilde{Z}^i\}_{i \in \mathbb Z_m}$
\STATE \#\#\# Preliminary steps -- subsection $B$
\FOR{$i \in \mathbb{Z}_m$ \textbf{in parallel}}
    \STATE $\mathcal{C}^i$ generates orthogonal matrix $Q$, using $\Xi$
    \STATE $\mathcal{C}^i$ generates orthogonal matrix $Q'$, using $\Xi+1$
    \STATE $\mathcal{C}^i$ generates invertible diagonal matrix $S$, using $\Xi$
    \STATE $\mathcal{C}^i$ calculates $M=QSQ'$
    \STATE $\mathcal{C}^i$ generates noise matrix $R^i$
\ENDFOR
\STATE \#\#\# Main process -- subsection $C$
\FOR{$i \in \mathbb{Z}_m$ \textbf{in parallel}}
    \STATE $\mathcal{C}^i$ calculates $\tilde{X}^i = X^i M$
    \STATE $\mathcal{C}^i$ sets $W^i=R^i$
    \FOR{$j \in \mathbb{Z}_{N^i}$}
        \STATE $\mathcal{C}^i$ sets $k={(\tilde{Z}^i)}_{j}$
        \STATE $\mathcal{C}^i$ calculates in-place ${(W^i)}_{k \cdot}= {(W^i)}_{k \cdot} + {(X^i)}_{j \cdot} $
    \ENDFOR
    \STATE $\mathcal{C}^i$ sends $R^i$ to $\mathcal{A}$ and $W^i$ to $\mathcal{P}$
\ENDFOR
\STATE $\mathcal{A}$ calculates $R=\sum_{i \in \mathbb{Z}_m} R^i$
\STATE $\mathcal{P}$ calculates $W=\sum_{i \in \mathbb{Z}_m} W^i$
\STATE $\mathcal{A}$ sends $R$ to $\mathcal{P}$
\STATE $\mathcal{P}$ calculates $X_{\text{masked}}=R-W$
\STATE \#\#\# Main process -- subsection $D$
\STATE $\mathcal{P}$ executes IF or EIF on $X_{\text{masked}}$
\STATE $\mathcal{P}$ communicates results to clients
\STATE Clients remove outliers from their data
\end{algorithmic}
\end{algorithm}

\subsection{Main process -- outlier detection and results communication}

The principal server $\mathcal{P}$ can now apply an outlier detection method, such as IF or EIF, to the masked data matrix. As an output of this step, the server obtains an outlier score for each data point involved in the process (even though, thanks to the permutation in Algorithm~\ref{algo:starting} it does not know to which client a given outlier score is associated).

If the consortium agrees, $\mathcal P$ could send the entire vector of scores to all clients, so that each client gains knowledge of the overall outlier landscape. The clients could then evaluate the outlier scores of their data points (since they know at which elements of the vector they are stored) and autonomously decide whether to discard them. Otherwise, in a more conservative approach, $\mathcal P$ could make the choice of which data points to qualify as outliers, so that they get discarded. It could then send to the clients only the equations of hyperplanes characterizing the regions of the ``masked space" where outliers lie. In this way, each client could check whether any of their points lies in such a region and, if so, discard them.

\section{Experiments} \label{sec:experiments}

\subsection{Design}

We tested our approach by conducting outlier detection on several different well-known datasets, which we downloaded from \cite{rayana2016}. These datasets include entries manually labeled as \textit{true} outliers and are commonly used as benchmarks for testing and comparing outlier detection methods. They were selected to provide a good variety of scenarios, including varying sizes, numbers of variables, and percentages of outliers. The main characteristics of the datasets are summarized in Table \ref{tab1}.

For each dataset we used the same procedure. Specifically, we compared executions of IF and EIF conducted without applying any transformation, which we will refer to as the standard approach, to executions of IF and EIF following the scheme presented in Section \ref{sec:methodology}, which we will refer to as the multiparty approach. We opt for the term ``multiparty" instead of ``federated", since our methodology does not build a global model through the aggregation of local contributions, which is a key aspect associated with the word ``federated", but builds a global model through pooled masked data. In the standard approach, we conducted 100 runs using different random seeds. In the multiparty approach, we tested four different values of the parameter $T$, which controls the scaling, and for each one we performed 100 different runs. In all the multiparty executions, we uniformly partitioned the data among three different clients. We did not explore situations where the data are non-iid among clients or where the local datasets have significantly different sizes, as the method is not sensitive to these aspects. Indeed,  it is equivalent for the method, for example, if outliers are held all by the same client or uniformly distributed among clients. This is because the masked data are pooled together for the outlier detection step. 

All simulations were conducted in R using the \texttt{isotree} package \cite{isotree}. We used $t=100$ trees, as suggested in \cite{chabchoub2022depth}, where the authors observed that this hyperparameter does not substantially affect the results. Conversely, in the same article, the authors observed that number of data points used to build each tree, $\psi$, has an impact on results, with not-too-large values of $\psi$ generally providing better results. We used $\psi=256$, which is a popular choice used by several publications (note that if a dataset has a number of data points $N<256$ then the function sets $\psi=N$).

\begin{table}[b]%[htbp]
\caption{Datasets for experiments}
\begin{center}
\begin{tabular}{|c|c|c|c|}
\hline
\textbf{Dataset} & \textbf{\textit{\# of points}}& \textit{\textbf{\# of variables}} & \textbf{\textit{\# (\%) of outliers}} \\
\hline
%\hline
Arrhythmia & 452 & 274 & 66 (15\%) \\
\hline
Cardio &  1831 & 21 & 176 (9.6\%)  \\
\hline
Glass & 214 & 9 & 9 (4.2\%) \\
\hline
Ionosphere & 351 & 33 & 126 (36\%)  \\
\hline
Lympho &  148 & 18 & 6 (4.1\%)  \\
\hline
Mammography &  11183 & 6 & 260 (2.32\%)  \\
\hline
Mnist &  7603 & 100 & 700 (9.2\%)  \\
\hline
Musk &  3062 & 166 & 97 (3.2\%) \\
\hline
Satellite &  6435 & 36 & 2036 (32\%)  \\
\hline
Shuttle &  49097 & 9 & 3511 (7\%)  \\
\hline
Speech &  3686 & 400 & 61 (1.65\%)  \\
\hline
Thyroid &  3772 & 6 & 93 (2.5\%)  \\
\hline
Vertebral &  240 & 6 & 30 (12.5\%)  \\
\hline
Vowels &  1456 & 12 & 50 (3.4\%)  \\
\hline
\end{tabular}
\label{tab1}
\end{center}
\end{table}

\subsection{Results}

\begin{figure}[htbp]
\centerline{\includegraphics[scale=.80]{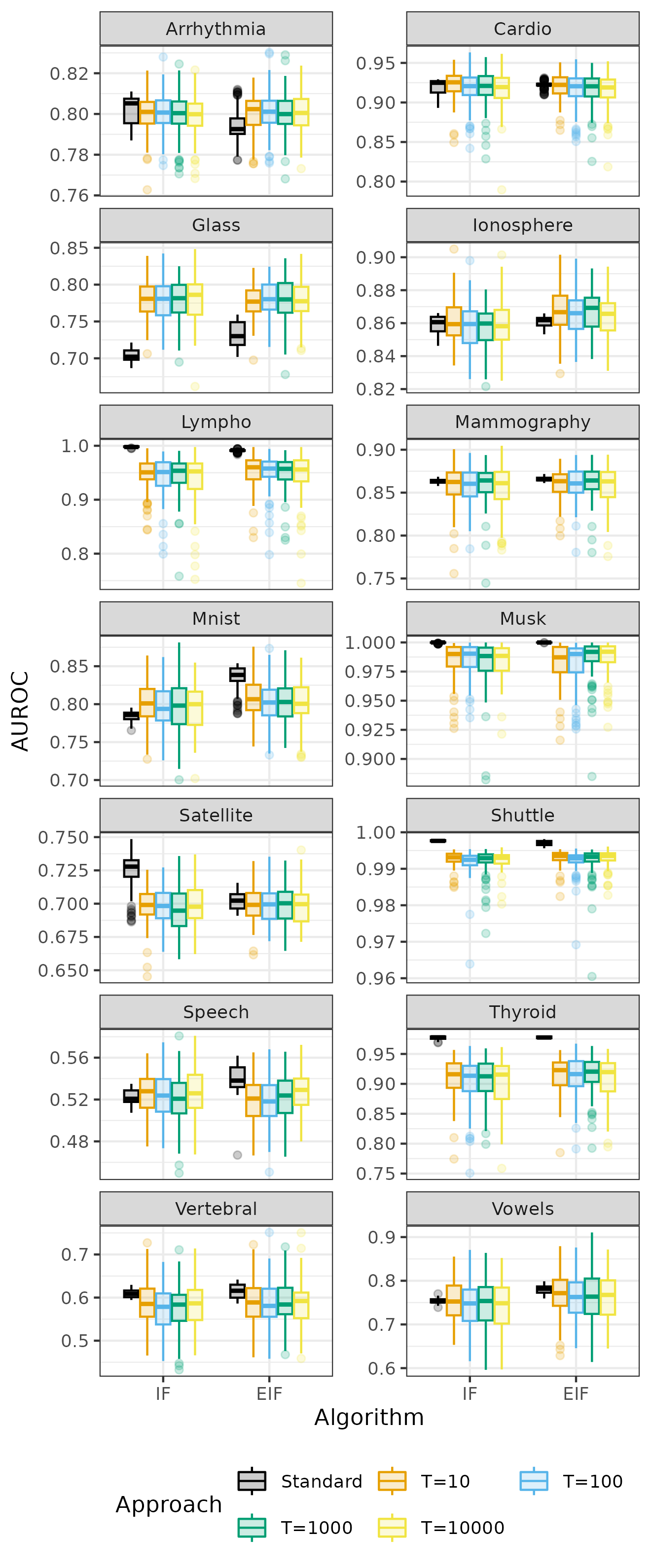}}
\caption{Performance for outlier classification in different datasets, for both IF and EIF. For each case, a reference approach and four multiparty approaches (corresponding to four different values of the parameter $T$) are shown. Each box in the boxplots includes 100 runs.}
\label{fig:plot}
\end{figure}

Since the datasets that we considered for evaluation contain labels identifying \textit{true} outliers, as is common for comparing outlier detection methods, we compared the performance of the different approaches in terms of AUROC. The results of this analysis are shown in the boxplot in Figure~\ref{fig:plot}, where each box represents the performance of 100 different executions for a given dataset, algorithm, hyperparameter choice, and approach (distinguishing the different choices of the parameter $T$ in the multiparty approach).

For all the different datasets and algorithms, the performance associated with the different choices of $T$ consistently does not exhibit any relevant difference. Moreover, for all datasets, the performance of the multipart approach using IF and EIF are equivalent. As we will comment in the next section, this is related to the effect of the masking on data.

By comparing the average performance of the standard and multiparty approaches in the different cases, it emerges that they are in general comparable. There are cases where they are totally equivalent (e.g., Cardio and Mammography), cases where the multiparty approach performs worse than the standard one (e.g., Lympho and Thyrod), and cases where the multiparty approach performs better than the standard one (e.g., Glass and IF Mnist). In all cases where the multiparty approach performs worse than the standard one, the difference is limited to a few percentage points. The Glass dataset was also analyzed in \cite{li2023federated}, with performance analogous to the standard IF approach, while our multiparty approach outperforms both standard IF and EIF. We did not compare with respect to other datasets analyzed in \cite{li2023federated}, as they were specific to the IoT domain, which is outside the scope for this work.

As a general observation, the boxplots characterizing the multiparty approach are generally wider than those characterizing the standard approach, showing more variability in the results. For this reason, a consortium using the method might find it useful to perform more than one execution, with different choices for the global seed $\Xi$. For example, in a scenario where the principal server communicates to the clients only the regions of the space associated with outliers, the consortium could execute it three times, and clients could discard the points that are qualified as outliers at least two out of the three times. This will contribute to making the results more robust. Note that the server would not be able to count how many times it has qualified a given data point as an outlier, as the same data point will be stored in different rows across executions.

\section{Discussion} \label{sec:discussion}

%In this section, we comment more in depth on several points of interest such as the multiplicative masking that we use, potential collusion scenarios among parties, and strategies for enhancing privacy in the approach.

\subsection{On the masking matrix $M$}

From a geometrical point of view, the masking matrix $M=QSQ'$ represents a composite transformation encoding two rotations and a scaling. However, this is not a special characteristic of $M$, as any real matrix can be decomposed into a product commonly denoted as $U \Sigma V^T$, where $U$ and $V$ are real orthogonal matrices and $\Sigma$ is a real non-negative diagonal matrix. This is, in fact, the Singular Value Decomposition (SVD) of the matrix, whose existence is guaranteed by an existence theorem. The reason why we decided to build the matrix $M$ as a product of matrices instead of randomly generating it element-wise is that, in our approach, we can generate it in a controlled manner. In particular, by choosing the elements of $S$ in the interval $(1, T)$, we can ensure that $M$ is far from singularity.
However, the QR decomposition required for generating the random orthogonal matrices %$Q$ and $Q'$ %in the local generation of $M$ 
is the operation with the most significant impact on the method's runtime, with a time complexity of \(O(n^3)\)\cite{mezzadri2006generate}.
%The local generation of $M$ is the operation with the most significant impact on the method's runtime due to the QR decomposition required for generating random orthogonal matrices, which has a time complexity of \(O(n^3)\) \cite{mezzadri2006generate}.

It is interesting to observe how the different elementary transformations composing the masking transformation contribute to obfuscating different aspects of the data (while not altering the isolation status of outliers). Anisotropic scaling alters the absolute and relative distances among data points, as well as densities, correlations, norms of data vectors, and the singular values of the data matrix (which rotation preserves), while rotation alters the rankings between data points with respect to a given axis of the coordinate system (which anisotropic scaling preserves). Moreover, the rotations and scaling operated by the masking transformation essentially generalize the split directions in IF, thus making its use substantially equivalent to the use of EIF, and explaining what observed in Figure~\ref{fig:plot} regarding their equivalence in the multiparty approach.

\subsection{Collusion among parties}

There are several possible collusion scenarios that we shall discuss, involving different kinds of parties and with different levels of criticality.

In the case where up to $m-2$ clients collude, they cannot univocally identify the owner of the data point associated with a given outlier score. However, if $m-1$ clients collude, they can trace which outlier scores belong to the $m$-th client.

If the auxiliary server colludes with one or more clients, the clients do not gain any additional information. In fact, thanks to the homomorphic encryption used in Algorithm \ref{algo:starting} for creating the starting points $\{s_i\}_{i \in \mathbb Z_m}$, the auxiliary server does not know the starting points of the different clients, and therefore it cannot associate an outlier score to the owner of the associated data point. Similarly, if the two servers collude, the principal server does not gain any additional information.

The most critical situation arises if the principal server colludes with one or more clients. In such a case, the colluding parties can reconstruct the entire data matrix. However, thanks to the use of homomorphic encryption for assigning starting points, they cannot determine to which specific client the individual data points belong. The assumption of a trusted principal server is therefore the most important one, and constitutes the main limitation of the method. However, there is a wide range of situations where this assumption can be considered reasonable (e.g., principal server belonging to a well-known and reputed academic institution, non-governmental organization or foundation).

\subsection{Strategies for further privacy} \label{subsec:smote}

To enhance privacy further, clients could implement strategies that modify their local datasets without compromising the identification of outliers. For instance, clients could execute IF locally to assess the extent to which their data points can be considered as local outliers. Subsequently, they could focus on data points with low outlier scores and employ strategies to modify them. For example, they might use downsampling, uniformly excluding data points with outlier scores below a given threshold. Alternatively, they could introduce new instances by leveraging techniques such as SMOTE \cite{chawla2002smote}. SMOTE can effectively generate synthetic instances that are interpolations of the original ones, thus adding data points without altering the isolation pattern of true outliers. A coupled use of IF and SMOTE has already been proposed in \cite{zheng2019improved}, even though with a different scope. Clients could even consider fabricating fake outliers to obfuscate the true number of outliers from the server. However, we caution against this approach due to the potential risks it poses to the integrity of the process.

\section{Conclusion and outlook} \label{sec:conclusion}

We have presented a technique for identifying global outliers in a Federated Learning setting using a two-server approach. The clients provide one of the two servers with masked data, where the masking preserves the ability to identify outliers while maintaining privacy. The server applies an outlier detection method, such as Isolation Forest or Extended Isolation Forest, and then communicates the results to the clients either by providing outlier scores or indicating regions of the masked space from which data points should be discarded. Our tests on various datasets show that the performance is comparable to traditional methods applied to unmasked data at a single site.

It would be interesting to test the proposed masking scheme with other outlier detection algorithms. We initially tested it with the Isolation Forest algorithm (and its extended version), as it seemed the most natural choice for a masking transformation enforcing a scaling and two rotations. While we expect the method not to provide satisfactory results when combined with density-based algorithms (as density is altered by the transformation), it could potentially yield good results with other classes of algorithms. Moreover, it would be interesting to evaluate whether the same masking scheme permits achieving other tasks, such as batch effect detection, as we expect the ``batchness" of data to be preserved by the transformation.

%\section*{Acknowledgment}
%...

\bibliography{IEEEabrv,bibliography}

\end{document}